\documentclass{l4dc2026}
\jmlrproceedings{}{}\jmlrvolume{}\jmlrpages{}\jmlryear{}

\usepackage{times}
\usepackage{booktabs}
\usepackage{multirow}
\usepackage{xcolor}

\title[Delay Randomization Widens the Delay Margin]{Delay Randomization Widens the Delay Margin of Learned Humanoid Motion Trackers}

\author{%
 \Name{Shaoqin Li$^{1}$ and Yi Sun$^{1,*}$}\\
 \addr $^{1}$Department of Electrical Engineering, The City College of New York, New York, NY 10031, USA\\
 \addr $^{*}$Corresponding author: {\color{black}\href{mailto:ysun@ccny.cuny.edu}{\color{black}ysun@ccny.cuny.edu}}%
}

\begin{document}

\maketitle
\thispagestyle{plain}
\pagestyle{plain}

\begin{abstract}%
Learned humanoid controllers are usually trained in one simulator and then checked in a second one before they go near hardware.
We ask what that check says about command delay.
We train motion-tracking policies for a 24-joint humanoid on five motions taken from monocular video and test them in an independent CPU MuJoCo harness, which we first verify against the training model.
Every policy completes its motion in at least 26 of 30 episodes under the training randomization.
With 40\,ms of constant command delay, however, survival drops to 15\% on average (0--45\%), and the delay margin is only 20--30\,ms.
Randomizing the delay during training (0--50\,ms) lifts survival at 40\,ms to 95\% on average (75--100\%) and moves the margin to 50--70\,ms.
A control arm trained for the same number of extra iterations without delay does not improve, so the gain comes from the randomization itself.
This holds for all five motions with three training seeds each, and a policy trained in PhysX behaves the same way.
The price is slightly looser tracking (0.02--0.06\,rad on four motions) and, for the most dynamic motion, less robustness to other perturbations.
We also show that the first sim-to-sim gap we saw came from the harness, a missing floor and a wrong timestep, and not from the physics engines.
The 0--20\,ms delay ranges common in locomotion training lie inside the margin an unrandomized tracker already has.
The benefit comes from randomizing beyond it, so the range should be set from the robot's measured latency, and delay margin belongs next to other sim-to-sim metrics.%
\end{abstract}


\section{Introduction}

Reinforcement-learned motion trackers can reproduce highly dynamic human motions on humanoid robots \citep{liao2025beyondmimic,he2025asap}.
They are trained in a GPU simulator with randomized dynamics \citep{tobin2017domain,peng2018sim} and, before hardware, commonly checked in a second simulator \citep{gu2024humanoid,lei2025polysim}.
The policy then runs on a robot whose control loop always has some latency between the policy output and the actuator response \citep{tan2018simtoreal,hwangbo2019learning}.
In classical control, delay is a first-order concern: it erodes phase margin and can destabilize an otherwise stable loop \citep{richard2003time}.
For learned controllers it is usually handled by randomizing a small delay during training; humanoid locomotion frameworks report ranges such as 0--20\,ms \citep{wang2025booster,seo2025learning}.
Few papers report how much delay a learned \emph{motion tracker} actually tolerates, or what randomizing it in training buys.

This paper measures it.
The setting is a 24-joint humanoid tracking five motions recovered from ordinary video.
Each policy is trained in mjlab (MuJoCo-Warp) \citep{zakka2026mjlab} with the BeyondMimic tracking recipe \citep{liao2025beyondmimic} and evaluated in an independent, CPU-based MuJoCo harness \citep{todorov2012mujoco}.
We make three contributions.
\begin{enumerate}
  \item \textbf{A verified evaluation protocol.} The harness rebuilds the training model from the robot description and is validated to reproduce training-time observations and outcomes. A one-at-a-time ablation shows that the first sim-to-sim ``gap'' we observed came entirely from two settings (a missing floor and a 2.5$\times$ smaller timestep), not from the physics engine (Section~\ref{sec:harness}).
  \item \textbf{Delay margins of learned trackers.} Survival as a function of constant command delay, at 10\,ms resolution, for policies trained with and without delay randomization, across five motions and three training seeds (Section~\ref{sec:margin}).
  \item \textbf{A controlled attribution.} A control arm with equal additional training but no delay isolates the effect of delay randomization from the effect of training longer (Section~\ref{sec:attr}). A PhysX-trained policy shows the same failure and the same remedy (Section~\ref{sec:cross}).
\end{enumerate}

\section{Related Work}

\paragraph{Delay in learned control.}
Simulating latency was one ingredient of early sim-to-real locomotion \citep{tan2018simtoreal}, and actuator dynamics, including delay, were later learned from data \citep{hwangbo2019learning}.
Current humanoid frameworks randomize actuation latency during training, typically up to 20\,ms \citep{wang2025booster,seo2025learning}.
In reinforcement learning, random delays have been treated as part of the decision process itself \citep{bouteiller2021reinforcement}.
Time-delay systems are a classical topic in control \citep{richard2003time}; our measurements can be read as an empirical delay margin of a learned nonlinear closed loop.

\paragraph{Motion tracking and sim-to-sim checks.}
Humanoid motion trackers learned from human motion now reach agile skills on hardware \citep{liao2025beyondmimic,he2025asap,allshire2025visual}.
Humanoid-Gym made sim-to-sim verification an explicit step \citep{gu2024humanoid}, and PolySim trains across several simulators at once \citep{lei2025polysim}.
ASAP corrects residual dynamics mismatch with real data \citep{he2025asap}.
We focus on a narrower question these works leave open: how delay tolerance of a tracker depends on delay randomization, measured with an evaluation harness whose own fidelity is checked first.

\section{Setup}
\label{sec:setup}

\paragraph{Robot and controller.}
The humanoid model has 24 actuated joints (six per leg, two in the waist, five per arm) and a mass of 28.9\,kg.
Each joint is driven by a PD loop at 200\,Hz (5\,ms physics step) with torque limits from the manufacturer's deployment description.
The policy runs at 50\,Hz and outputs joint-position targets as scaled offsets from a default standing pose.
It observes a 135-dimensional vector: the reference joint positions and velocities, the torso position and orientation relative to the reference, base linear and angular velocity, joint positions and velocities, and the previous action.
Policies are trained with PPO \citep{schulman2017proximal} on 4,096 parallel robots.
Training randomizes observation noise, initial pose and velocity, velocity pushes every 1--3\,s, torso center-of-mass offset, joint-encoder bias ($\pm$0.01\,rad) and foot friction (0.3--1.2).

\paragraph{Delay model.}
The \emph{delay-randomized} variant adds a command delay on the actuator targets: each robot draws an integer lag $L\sim\mathcal{U}\{0,\dots,10\}$ physics steps (0--50\,ms), redrawn every 200 physics steps (1\,s) with a per-robot phase, and zero after a reset until the first redraw.
A target issued at physics step $k$ is applied at step $k+L$.

\paragraph{Motions.}
Five reference motions come from monocular video through GVHMR \citep{shen2024gvhmr} and GMR \citep{araujo2025retargeting}, retargeted at each video's native frame rate (Table~\ref{tab:motions}).
They differ in camera (fixed or moving), duration (8.7--26.4\,s) and character (a tennis serve, three dances, shadow boxing).

\begin{table}[t]
\centering\small
\begin{tabular}{lllcc}
\toprule
Motion & Source & Camera & Duration & Video fps \\
\midrule
Tennis serve & GVHMR example clip & fixed & 10.4\,s & 29.8 \\
Dance (studio) & stock video & fixed & 26.4\,s & 25 \\
Dance (home) & stock video & fixed & 8.7\,s & 30 \\
Shadow boxing & stock video & moving & 15.0\,s & 24 \\
Street dance & stock video & moving & 15.0\,s & 25 \\
\bottomrule
\end{tabular}
\caption{Reference motions, all recovered from monocular video (GVHMR $\rightarrow$ GMR).}
\label{tab:motions}
\end{table}

\paragraph{Training schedule.}
For each motion we train a base policy without delay (3,000 iterations; 12,998 for tennis).
From that checkpoint we continue for 3,000 iterations in two arms, each with three training seeds: \emph{control} (same task, no delay) and \emph{delay} (delay-randomized task).
The control arm is what isolates the delay randomization from the additional training.

\paragraph{Evaluation harness.}
Policies are exported to ONNX and run in plain MuJoCo on the CPU.
The harness builds the physics model from the robot description with MuJoCo's specification API, reproducing the training model's floor, contact parameters, 5\,ms timestep, solver settings and implicit-fast integrator.
It reimplements the observation, the training randomization and the training termination rules (torso orientation, torso height or end-effector height deviating beyond 0.25\,m).
An episode \emph{survives} if it completes the whole motion.
Stress tests add, on top of the training randomization, a \emph{constant} command delay $d\in\{0,10,\dots,80\}$\,ms applied at physics-step resolution, $3\times$ pushes, foot friction 0.15, $\pm$20\% PD gains, a 2\,kg payload, or the removal of base velocity from the observation.
Each delay level uses 20 episodes; the plain training randomization uses 30.

\section{Results}

\subsection{Verifying the harness first}
\label{sec:harness}

The first version of the harness loaded the robot description directly, and the tennis policy diverged (mean joint error 4.6--8.9\,rad, against 0.78 inside training).
Running the network on GPU in PyTorch instead of ONNX on CPU changed nothing.
Restoring the training model's settings one at a time located the cause (Table~\ref{tab:ablation}): the robot description has no ground plane, and the harness stepped physics at 2\,ms instead of 5\,ms.
Contact dimension, solver iterations, friction and contact priority each moved the error by less than 0.006\,rad.
With the model matched, the harness reproduces training: the policy tracks the whole clip, and for the PhysX-trained policy the harness's first observation equals the training simulator's to $10^{-5}$.
The harness is deterministic: re-running a sweep from a fresh installation reproduced the recorded results episode for episode (checked for the tennis delay-randomized policy and the PhysX policy).

\begin{table}[t]
\centering\small
\begin{tabular}{lc}
\toprule
Raw-description setting kept & Joint error (rad) \\
\midrule
none (training-equivalent model) & 0.577 \\
contact dimension / solver iterations / contact priority & 0.576 / 0.577 / 0.577 \\
foot friction & 0.582 \\
all four above & 0.582 \\
\textbf{timestep (2\,ms)} & \textbf{1.719} \\
\textbf{no floor} & \textbf{6.969} \\
\bottomrule
\end{tabular}
\caption{One-at-a-time ablation, tennis policy, nominal rollout. Joint error is the L2 norm of the 24-joint error vector averaged over the episode.}
\label{tab:ablation}
\end{table}

\subsection{Delay margin with and without delay randomization}
\label{sec:margin}

\begin{figure}[t]
\centering
\includegraphics[width=\linewidth]{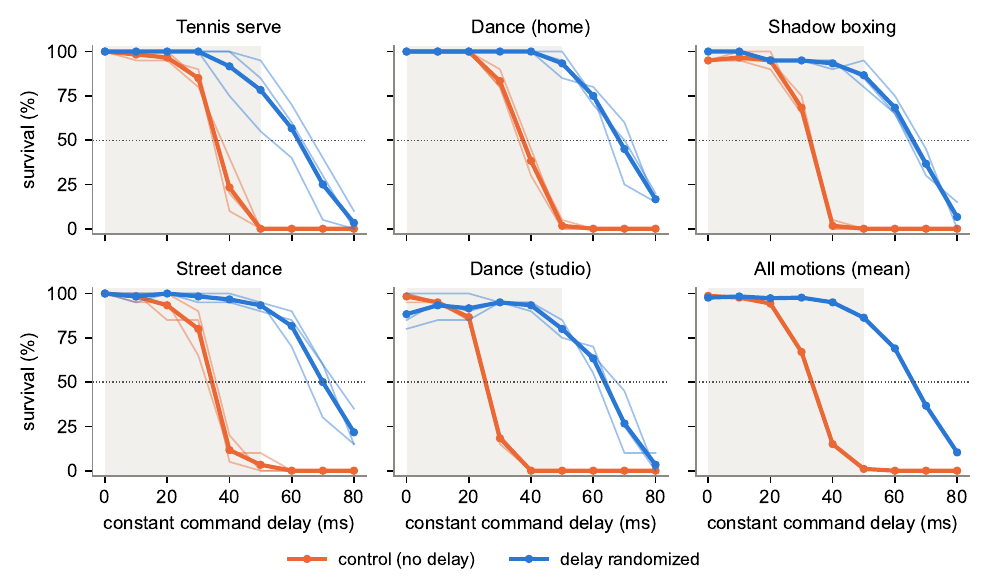}
\caption{Survival against constant command delay in the independent MuJoCo harness (20 episodes per point). Thin lines are individual training seeds, thick lines their mean; orange: control (additional training, no delay), blue: delay-randomized training. The grey band is the range used in training (0--50\,ms). The last panel averages all fifteen policies per arm.}
\label{fig:margin}
\end{figure}

Figure~\ref{fig:margin} shows survival against constant command delay.
We define the \emph{delay margin} of a policy as the largest delay on our 10\,ms grid at which at least half of the episodes survive, with all smaller delays also at or above half.
Without delay randomization (control arm) the margin is 30\,ms for four motions and 20\,ms for the studio dance, identical for all three training seeds of each motion.
The drop is a cliff: survival averages 67\% at 30\,ms, 15\% at 40\,ms, 1\% at 50\,ms and 0\% from 60\,ms on.
The delay-randomized arm has a margin of 50--70\,ms (60\,ms for 10 of 15 policies, 70\,ms for four; 14 of 15 exceed the 50\,ms maximum used in training).
It averages 95\% survival at 40\,ms, 86\% at 50\,ms, 69\% at 60\,ms and 37\% at 70\,ms, and degrades gradually instead of abruptly.

\subsection{Is it the delay or the extra training?}
\label{sec:attr}

Table~\ref{tab:attr} compares the two arms with the base policy, which is the common starting point of both.
At 40\,ms every delay-randomized policy survives at least 15 of 20 episodes and every control policy at most 9; the ranges of the two arms do not overlap for any motion, and at 60\,ms the control arm never survives.
The base policies behave like the controls (0--2 of 20 at 40\,ms), so additional training without delay does not help.
The one visible effect of extra training is on the studio dance, which is the hardest motion to track: its joint error falls from 2.27 to 1.88\,rad and its margin rises from 10\,ms (8 of 20 episodes survive at 20\,ms) to 20\,ms.
The delay arm shows that the order of magnitude of the margin is set by the delay range in training.

\begin{table}[t]
\centering\small
\begin{tabular}{lccccc c}
\toprule
 & \multicolumn{3}{c}{40\,ms} & \multicolumn{3}{c}{60\,ms}\\
\cmidrule(lr){2-4}\cmidrule(lr){5-7}
Motion & base & control & delay & base & control & delay \\
\midrule
Tennis serve & 2 & 4.7 (2--8) & 18.3 (15--20) & 0 & 0 & 11.3 (8--14) \\
Dance (home) & 0 & 7.7 (6--9) & 20 & 0 & 0 & 15.0 (14--16) \\
Shadow boxing & 0 & 0.3 (0--1) & 18.7 (18--19) & 0 & 0 & 13.7 (13--15) \\
Street dance & 1 & 2.3 (1--4) & 19.3 (19--20) & 0 & 0 & 16.3 (14--18) \\
Dance (studio) & 0 & 0 & 18.7 (18--19) & 0 & 0 & 12.7 (11--14) \\
\bottomrule
\end{tabular}
\caption{Episodes survived out of 20 at constant command delay. Control and delay: mean over three training seeds (range); base: single policy.}
\label{tab:attr}
\end{table}

\subsection{Other perturbations and costs}

Table~\ref{tab:cost} reports the cost of delay randomization.
Nominal tracking error, averaged over surviving episodes, is slightly higher for the delay arm on four of five motions (by 0.02--0.06\,rad), and lower for the studio dance.
Robustness to the other perturbations is mixed rather than uniformly better or worse.
Delay randomization roughly doubles survival at foot friction 0.15 for the tennis serve (23\% to 48\%) and is slightly higher for the two moving-camera motions, while 3$\times$ pushes, $-20\%$ PD gains and a missing base velocity change survival by 10 points or less for four of the five motions.
The studio dance is the exception: the delay arm loses robustness to a 2\,kg payload (98\% to 82\%), pushes (10\% to 3\%) and, most strongly, to a missing base-velocity observation (37\% to 5\%), and two of its three seeds complete only 26 and 27 of 30 training-randomization episodes.
It is the fastest motion (joint error 1.8\,rad, three times the others), so a plausible reading is that it has little control authority to spare once part of it is spent on tolerating delay; we have not tested this.

\begin{table}[t]
\centering\footnotesize
\begin{tabular}{lccccccc}
\toprule
 & \multicolumn{2}{c}{Nominal} & \multicolumn{5}{c}{Survival (\%) under additional perturbation} \\
\cmidrule(lr){2-3}\cmidrule(lr){4-8}
Motion & error (rad) & surv.\ (\%) & push 3$\times$ & fric.\ 0.15 & $K_p{-}20\%$ & 2\,kg & no $v_\mathrm{base}$ \\
\midrule
Tennis serve & 0.60 / 0.65 & 100 / 100 & 63 / 63 & 23 / 48 & 95 / 90 & 98 / 100 & 40 / 30 \\
Dance (home) & 0.61 / 0.63 & 100 / 100 & 90 / 85 & 73 / 77 & 98 / 98 & 100 / 100 & 73 / 70 \\
Shadow boxing & 0.69 / 0.75 & 97 / 100 & 42 / 35 & 13 / 20 & 92 / 95 & 97 / 100 & 82 / 83 \\
Street dance & 0.59 / 0.61 & 99 / 99 & 25 / 33 & 73 / 83 & 88 / 93 & 100 / 98 & 63 / 65 \\
Dance (studio) & 1.88 / 1.82 & 97 / 92 & 10 / 3 & 57 / 52 & 85 / 75 & 98 / 82 & 37 / 5 \\
\midrule
Mean & -- & 98 / 98 & 46 / 44 & 48 / 56 & 92 / 90 & 99 / 96 & 59 / 51 \\
\bottomrule
\end{tabular}
\caption{Each cell is control / delay-randomized, averaged over three training seeds (20 episodes per seed and perturbation, 30 for the nominal survival). Nominal error is the L2 norm of the 24-joint error, averaged over surviving episodes.}
\label{tab:cost}
\end{table}

\subsection{A second physics engine}
\label{sec:cross}

Policies trained in Isaac Lab (PhysX) with the BeyondMimic code and the same humanoid task show the same pattern in the same harness, for one motion and one training run per arm.
The dance policy trained without delay survives 29 of 30 training-randomization episodes but 0 of 20 at 40\,ms.
The same task with delay randomization (same 0--50\,ms scheme, trained from scratch) survives 15 of 20 at 40\,ms, but only 17 of 30 under the plain training randomization, a regression that we did not investigate.
The reference for these two policies was an earlier version of the dance that ran 1.2$\times$ too fast (a frame-rate error we later corrected), so their numbers are not comparable with the mjlab dance above; the qualitative result is what we use.

\section{Discussion}

A standard synchronous sim-to-sim check would pass all of these policies: each completes its motion under the training randomization, and nearly all do at 0--10\,ms of delay.
The failure shows up only when delay is injected, and then it happens within one 10\,ms step (Figure~\ref{fig:margin}), so a check at one delay value can miss it.
Without delay randomization, this family of trackers tolerates 20--30\,ms.
The 0--20\,ms ranges common in locomotion training therefore sit inside a margin the policies already have.
The gain in our experiments came from randomizing well beyond that, up to 50\,ms, which produced margins of 50--70\,ms.
For a real robot this suggests measuring the command latency (policy output to actuator response), randomizing over a range that covers it with room to spare, and reporting the delay margin with the other sim-to-sim numbers.
Measuring the margin is cheap, because the evaluation harness only needs a delay buffer.

For most motions the price is small, 0.02--0.06\,rad of nominal tracking error.
It is larger for the most demanding motion, where robustness to other perturbations also dropped.
We have not tried a wider range, a curriculum, or giving the policy the current delay as an input, and any of them might avoid this.
These margins are the empirical analogue of the delay margin of a linear loop \citep{richard2003time}, but we claim no quantitative link: the closed loop is nonlinear, the policy is not told the delay, and we did not examine why the unrandomized policies fail near 35\,ms.

\paragraph{Limitations.}
All results are in simulation, with idealized actuators (PD with torque limits).
One robot, five motions recovered from third-party video, three training seeds per arm for the mjlab policies and a single run for the PhysX policy.
The seeds of an arm continue from the same base checkpoint, so the seed spread understates the variation that retraining from scratch would add.
Survival is binary and measured over 20 episodes per delay level.
The harness holds delay constant for an episode, which is stricter than training, where the lag is redrawn every second and starts at zero.
The physical robot's command latency has not yet been measured; that measurement, and deployment, are our next steps.

\section{Conclusion}

We measured how much constant command delay learned humanoid motion trackers tolerate, using an independent simulator that we checked against the training model first.
Trained without delay, policies for five video-derived motions tolerate only 20--30\,ms, and survival falls from two thirds to under 15\% in a single 10\,ms step.
Randomizing delay up to 50\,ms in training moves the margin to 50--70\,ms for every motion and seed, and a matched control shows that longer training alone does not do this.
The cost is small for most motions and larger for the most dynamic one.
The next steps are to measure the robot's actual command latency and to test whether the simulated margin predicts behavior on the robot.

\bibliography{refs}

\end{document}